\documentclass[runningheads]{llncs}
\usepackage[T1]{fontenc}
\usepackage{graphicx}
\usepackage{booktabs}
\usepackage[misc]{ifsym}

\usepackage{mwe}
\usepackage{algorithm}
\usepackage{algorithmic}
\usepackage{times}
\usepackage{makecell}
\usepackage{amsmath}
\usepackage{amssymb}
\usepackage{soul,xcolor}
\usepackage{siunitx}
\usepackage{url}
\usepackage{multirow}
\usepackage[hidelinks]{hyperref}
\usepackage{tablefootnote}
\usepackage[autostyle]{csquotes}

\begin{document}

\title{{\em Hi Model, generating \enquote{nice} instead of \enquote{good} is not as bad as generating \enquote{rice}!} Towards Context and Semantic Infused Dialogue Generation Loss Function}

\titlerunning{Towards Context and Semantic Infused Dialogue Generation Loss Function}

\author{Abhisek Tiwari$^1$, Muhammed Sinan$^1$, Kaushik Roy$^2$, Amit Sheth$^2$, Sriparna Saha$^1$ \\ {\and Pushpak Bhattacharyya$^3$}}

\authorrunning{ }

\institute{$^1$Indian Institute of Technology Patna, India \\ 
        $^2$Artificial Intelligence Institute, University of South Carolina, USA \\
        $^3$Indian Institute of Technology Bombay, India}

\maketitle              

\begin{abstract}
Over the past two decades, dialogue modeling has made significant strides, moving from simple rule-based responses to personalized and persuasive response generation. However, despite these advancements, the objective functions and evaluation metrics for dialogue generation have remained stagnant. These lexical-based metrics, e.g., cross-entropy and BLEU, have two key limitations: (a) {\em word-to-word matching without semantic consideration:} It assigns the same credit for failure to generate \enquote{nice} and \enquote{rice} for \enquote{good}, (b) {\em missing context attribute for evaluating the generated response:} Even if a generated response is relevant to the ongoing dialogue context, it may still be penalized for not matching the gold utterance provided in the corpus. In this paper, we first investigate these limitations comprehensively and propose a new loss function called Semantic Infused Contextualized diaLogue ({\em SemTextualLogue}) loss function. We also formulate an evaluation metric called {\em Dialuation}, incorporating both context and semantic relevance. We experimented with both non-pretrained and pre-trained models on two dialogue corpora, encompassing task-oriented and open-domain scenarios. We found that the dialogue generation models trained with {\em SemTextualLogue} loss attained superior performance compared to the traditional cross-entropy loss function. The findings establish that the effective training of a dialogue generation model hinges significantly on incorporating semantics and context. This pattern is also mirrored in the introduced {\em Dialuation} metric, where the consideration of both context and semantics correlates more strongly with human evaluation compared to traditional metrics\footnote{The code and dataset are available at \url{https://github.com/NLP-RL/SemTextualLogue-Loss}}.

\end{abstract}

\section{Introduction}
Building a human-like conversational agent has always been one of the primary goals of artificial intelligence \cite{allen2001toward}. Initially designed to aid humans, it has now evolved to such a degree that it is even being employed for casual conversation to fulfill the human desire for social interaction. Transitioning from rule-based ELIZA to advanced chatbots like Alexa and ChatGPT vividly illustrates the imperative of actively constructing proficient dialogue assistants \cite{ni2023recent}. The primary expectation from an adequate dialogue assistant is to provide an appropriate and contextually relevant response \cite{valizadeh2022ai}. To meet the expectation of an effective dialogue assistant, the use of a proper loss function and evaluation metric is crucial. These components essentially serve as the backbone and essence of a learning framework.

\begin{figure*}[hbt!]
    \centering
    \includegraphics[width=\linewidth]{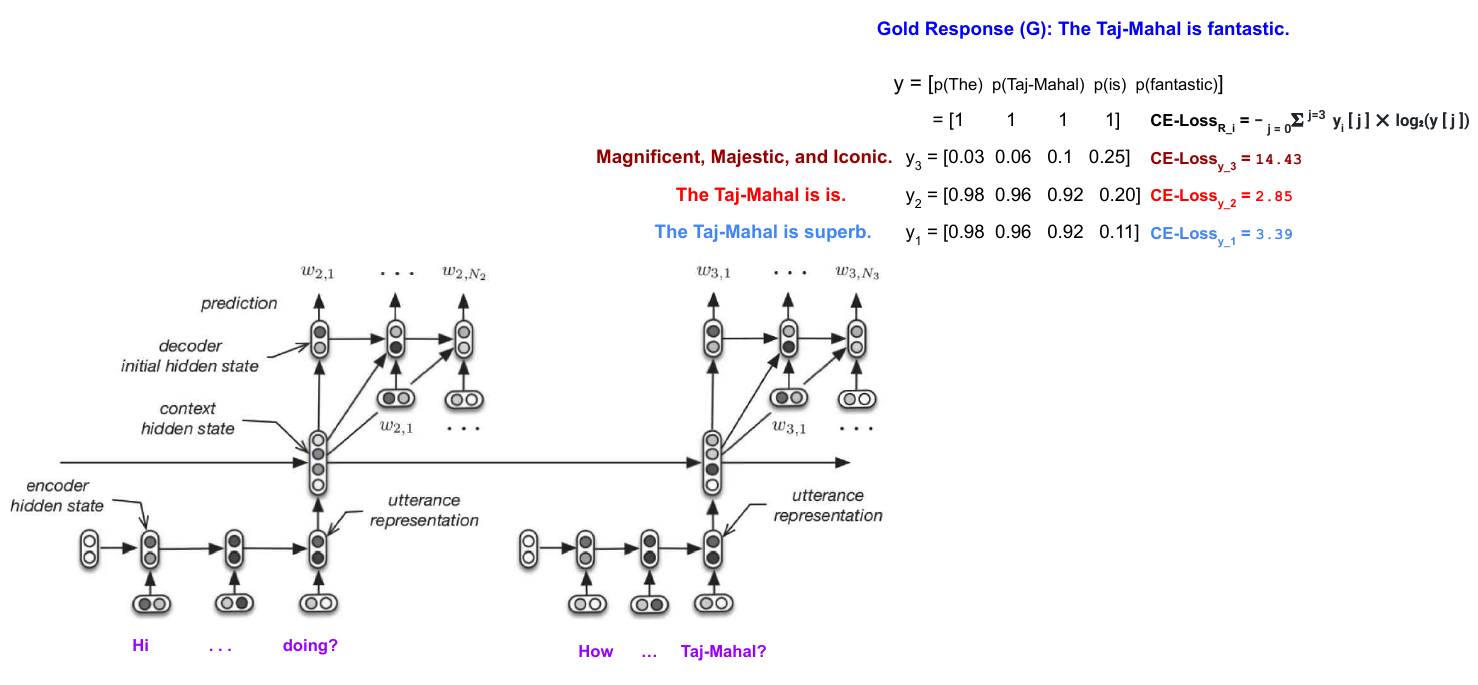}
    \vspace{-1.2em}
    \caption{Illustration of the key limitation of cross entropy for dialogue generation. Some adequate responses ($y_1$ and $y_2$) are equally or more penalized as useless response ($y_2$)}
    \label{CEL}
    \vspace{-0.5em}
\end{figure*}

 The most widely employed dialogue generation loss function is cross entropy (CE). The CE loss used in dialogue generation was borrowed from machine translation (MT) with the belief that the two tasks are identical. However, there are some substantial differences between the two tasks \cite{hu2020translation}: MT does not consider context, whereas it is a crucial aspect of dialogue generation. Furthermore, MT emphasizes lexical-based matching of generated text with reference text. The limitations caused by the discrepancies between the tasks are demonstrated in Figure \ref{CEL}. The generation model with CE loss has a fixed output expectation, which means that even semantically relevant responses ($y_1$) are being unfairly punished to the same extent or even more so than a meaningless response ($y_2$). The loss for the third response ($y_3$) is much higher because it has no uni-gram matches with the ground truth response. However, from a human perspective, the generated response appears to be contextually relevant and aligned with the reference response.

Our key objective from an artificial intelligence model is to replicate the process of human learning. Since humans are the ultimate consumers and evaluators of these models, their perception of learning and evaluation is crucial. Many dialogue generation works have recently discovered that word-based evaluation metrics do not strongly align with human judgment \cite{sato2020evaluating}. Humans consider a response appropriate and relevant if it conveys a similar meaning as expected in the context rather than a word-to-word match. Inspired by these observations, we investigate the importance of semantic-based evaluation and context relevance for dialogue loss and evaluation functions. Consequently, we develop a new dialogue generation loss function that incorporates semantic and contextual appropriateness in addition to lexical matching. Furthermore, we formulate a new context-infused, semantic-aware dialogue generation evaluation metric to validate the effectiveness of the loss function and assess its correlation with human judgment. 

\hspace{-0.53cm}\textbf{Research Questions} The paper aims to investigate the following three research questions related to dialogue generation: \textbf{(i)} Can the addition of a semantic-based evaluation component to the lexical-based loss function provide more accurate feedback on generated responses and thus improve the overall quality of dialogue generation? \textbf{(ii)} Can incorporating context relevance evaluation in the loss function improve the model's ability to generate responses that are more appropriate and coherent to the discourse? \textbf{(iii)} Will integrating the semantic component into the lexical-based evaluation metrics in dialogue generation result in a better correlation with human judgment? \\

\hspace{-0.53cm}\textbf{Key Contributions} The key contributions can be enumerated as follows: 
\begin{itemize}
    \item We thoroughly examine, analyze, and present some of the major drawbacks of the existing dialogue loss functions and evaluation metrics. 
    \item Inspired by human judgment, we propose a new dialogue generation loss function called {\em SemTextualLogue} loss that incorporates semantic and contextual appropriateness in addition to lexical-based divergence measure.
    \item We formulate a new {\em dia}logue generation eva{\em luation} metric named {\em Dialution}, which incorporates semantic similarity and contextual relevance. 
    \item The proposed loss function outperforms traditional cross-entropy loss significantly across various evaluation metrics on both the dialogue corpora. Furthermore, the evaluation metric was more related to human judgment than the existing metrics, such as BLEU and ROUGE. 
\end{itemize}

\section{Related Works}   The proposed work is relevant to the following three research areas: Dialogue generation, Dialogue loss functions, and Dialogue generation evaluation metrics. In the following paragraphs, we have summarized the relevant works and highlighted the research gap.

\hspace{-0.53cm}\textbf{Dialogue Generation} Dialogue generation can be approached using two primary methods: modular \cite{griol2008statistical}, and end-to-end \cite{serban2016building}. The latter approach, end-to-end dialogue modeling, has gained popularity in recent years as a result of the modular approach's high demand for annotated data. In the last few years, there have been three kinds of works carried out: knowledge-grounded dialogue generation \cite{zhao2020knowledge}, transfer-learning-based dialogue generation \cite{golovanov2019large}, and multimodal dialogue generation \cite{shen2021text}. In \cite{li2017adversarial}, the authors build a generative adversarial network (GAN) based dialogue generation framework. The framework involves a sequence-to-sequence model serving as the generator module and a reinforcement learning model acting as the discriminator. The generator generates responses, while the discriminator evaluates the distinguishability of the generated responses from the corpus and provides feedback to the generator accordingly.\\
\hspace{-0.53cm}\textbf{Dialogue Generation Loss Functions} Dialogue generation can be approached using two primary methods: modular \cite{griol2008statistical}, and end-to-end \cite{serban2016building}. The latter approach, end-to-end dialogue modeling, has gained popularity in recent years as a result of the modular approach's high demand for annotated data. We have summarized the utilized dialogue generation loss functions and their characteristics in Table \ref{LF}. In \cite{kovaleva2018similarity}, the authors have also considered generated words' semantic similarity with words of gold response in addition to word-to-word matching to mitigate the fixed target issue. However, it's worth noting that although they incorporated word-to-word semantics, there could be cases where different word arrangements have nearly identical meanings, such as with the phrases {\em Nice to see you} and {\em I am happy to meet you}. The CE loss favors maximum likelihood, and thus it suffers from a lack of diversity. To tackle this problem, the researchers \cite{ueyama2020diverse} devised an inverse n-gram frequency (INF) loss function, which is a weighted cross-entropy function based on n-gram frequency calculated from the entire corpus context. The INF loss function assigns weights to n-gram mismatches based on the inverse of their frequency, giving rare tokens more weight. This weighting mechanism results in more diverse responses, effectively addressing the issue of low diversity. 

\begin{table}[hbt!]
    \centering
     \vspace{-0.6em}
    \caption{Existing most employed dialogue generation loss functions and their characteristics}
    \scalebox{0.9}{
    \begin{tabular}{llccc}
    \hline
    \textbf{loss function}  & \textbf{Methodology} & \textbf{Context} & \textbf{Semantic} & \textbf{World Knowledge}\\   \hline
    Cross Entropy \cite{de2005tutorial} & word probability distribution divergence & $\times$ & $\times$  & $\times$ \\
    FOCAL loss \cite{wang2021diversifying}  & token frequency aware cross entropy  &  $\times$ & $\times$ & $\times$\\
    ITF loss \cite{ueyama2020diverse} & token frequency aware cross entropy  & $\times$ &$\times$ &$\times$ \\
    Inverse N-gram \cite{nakamura2019another}  & n-gram frequency aware cross entropy & $\times$ & $\times$ & $\times$\\
    FACE \cite{jiang2019improving}  & dynamic frequency aware cross entropy & $\times$ & $\times$ & $\times$ \\
    SBR \cite{kovaleva2018similarity}  & CE loss with word wise semantic similarity  & $\checkmark$ & $\times$ &  $\times$\\
    \hline
    \end{tabular}}
    \label{LF}
    \vspace{-1em}
\end{table}

\hspace{-0.53cm}\textbf{Dialogue Generation Evaluation Metrics} There are mainly two kinds of evaluation: automatic and human. All the existing popular automatic dialogue generation evaluation metrics are described in Table \ref{EM}. Despite dialogue being a contextual phenomenon, none of the metrics consider dialogue context for judging the relevance of the generated text. Consequently, many recent dialogue generation works and surveys \cite{sato2020evaluating} on dialogue generation have reported a poor correlation between these metrics and human judgment. In a few works \cite{conley2018improving}, the authors have considered the word-net-based distance between generated text and gold response to mitigate the word-to-word match constraint of the cross entropy loss. In \cite{zhang2019bertscore}, the authors proposed an embedding-based BERT semantic similarity evaluation metric, which computes cosine similarity between the BERT embeddings of generated text and expected response. In a recent study on natural language generation \cite{sellam2020bleurt}, a novel evaluation metric, BLEURT, was introduced. This metric employs a trained regression model to forecast the relevance between two sentences. While it implicitly encompasses semantic understanding due to extensive training, it falls short in accounting for the contextual nuances of dialogue \cite{feng2021survey}.

\begin{table}[hbt!]
    \centering
    \vspace{-0.6em}
    \caption{Evaluation metrics commonly used in dialogue generation along with their associated limitations}
    \scalebox{0.92}{
    \begin{tabular}{llccc}
    \hline
    \textbf{Evaluation Metric}    & \textbf{Methodology} & \textbf{Context} & \textbf{Semantic} & \textbf{World Knowledge}\\  \hline
     BLEU  \cite{papineni2002bleu} & n-gram overlap & $\times$ & $\times$  & $\times$ \\
     ROUGE  \cite{lin2004rouge}  & n-gram overlap  & $\times$ & $\times$ & $\times$\\
     METEOR \cite{banerjee2005meteor}  &  n-gram overlap  & $\times$ & $\times$ & $\times$\\
     BERT Similarity \cite{zhang2019bertscore}  & embedding-based similarity & $\times$&  $\checkmark$ & $\times$ \\
     Sentiment Distance \cite{conley2018improving} & sentiment coherence & $\times$& $\times$ & $\times$ \\
     Jaccard Similarity \cite{niwattanakul2013using}  &  n-gram overlap &  $\times$ & $\times$ & $\times$\\
     SynSet Distance \cite{conley2018improving} &  semantic knowledge-base distance &  $\times$  & $\checkmark$  & $\checkmark$ \\
     BLEURT \cite{sellam2020bleurt} & pre-trained LLM  &   $\times$ & $\checkmark$  &  $\times$ \\
     \hline
    \end{tabular}}
    \vspace{-2em}
    \label{EM}
\end{table}

\section{Proposed Methodology}
\begin{figure*}[hbt!]
    \centering
    \includegraphics[width=\linewidth]{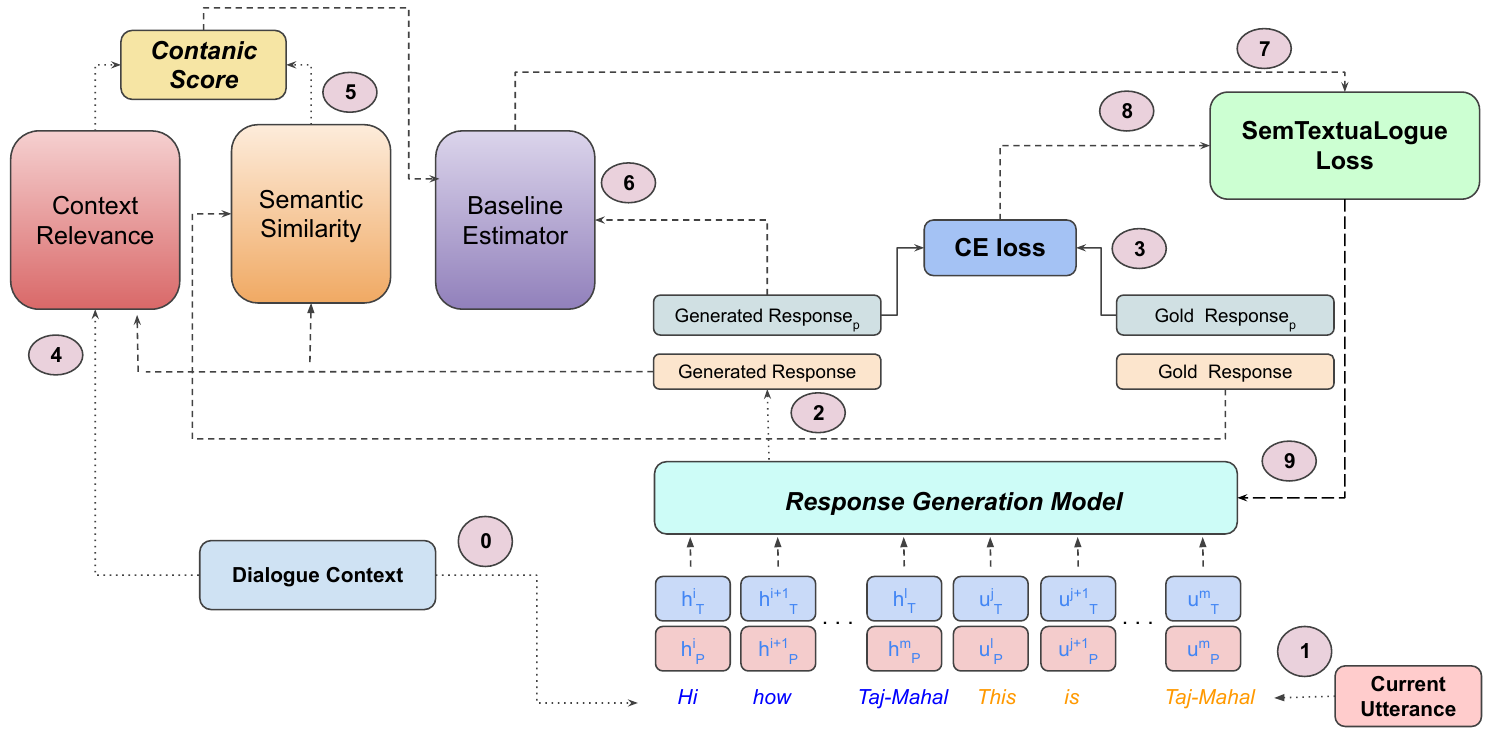}
    \vspace{-0.7em}
    \caption{Proposed architecture of semantic and context-reinforced dialogue generation. The dialogue generation model first generates a response based on dialogue context and current utterance. Subsequently, it calculates context and semantic relevance score ({\em Contanic}) and reinforces the feedback with the traditional cross-entropy loss}
    \vspace{-2em}
    \label{PM}
\end{figure*}

Dialogue generation involves producing a response within a conversation based on the dialogue context, including the current utterance. The model takes the dialogue context along with the current utterance as input and generates a response accordingly. The novelty in our approach lies in calculating the loss between the generated response and its gold counterpart, which has been described in subsequent sections. The proposed semantic and context-infused loss function incorporated dialogue generation model is illustrated in Figure \ref{PM}. It contains the following sub-modules: Response Generation Model, {\em Contanic}, and {\em SemTextualLogue} loss. The working of each of the sub-modules is explained and illustrated in the following sub-sections. We experiment with both types of dialogue generation models: non-pretrained models (transformer-based encoder-decoder) and pre-trained models such as GPT \cite{radford2019language} and LLaMA \cite{touvron2023llama}. Finally, the formulation of our proposed semantic and context-guided dialogue evaluation metric called {\em Dialution} is illustrated.

\subsection{Response Generation Model} The module takes dialogue context and current utterance as input and generates an output given the context. We have experimented with both kinds of dialogue generation models: encoder-decoder and decoder-only models. The encoder-decoder dialogue generation model employs two components: an encoder processes input context, like conversation history, while a decoder generates responses based on encoded context. In contrast, the decoder-only model generates responses directly from input context, often proving computationally efficient with competitive performance in dialogue tasks. The input sequence, consisting of both context and the current utterance, undergoes initial processing by being tokenized into a sequence of tokens. Each token is then transformed into a vector representation using an embedding layer. Subsequently, these embedded vectors are fed into the model's recurrent or transformer layers, which process them to generate a hidden state ($h_e$). This hidden state serves as the contextual information encoded from the input sequence. Using this processed hidden state, the model generates a token at each time step based on the probability distribution defined by:
\begin{equation}
\hat{y}_t = arg max_i P(V_i | y_1, y_2, ..., y_{t-1}, h_e)
\end{equation}
Here, $V$ denotes the vocabulary space, and $\hat{y}_t$ represents the generated token at the $t^{th}$ time step. Concatenating all generated tokens yields the final output sequence ($\hat{y}$). In conventional dialogue generation models, the comparison between the probability distribution of predicted tokens and that of the gold response is typically conducted using cross-entropy, and subsequently backpropagated. However, in order to address contextual and semantic appropriateness, we also integrate a {\em contanic} score into the loss component, which is then backpropagated, enabling the models to adapt to dialogue context and semantic measures and learn accordingly.

\subsection{{\em Contanic} Score}
In traditional dialogue generation, the predicted probability distribution is compared with the actual output sequence, and entropy deviation is calculated. The deviation is back-propagated to the network, and the parameter gets adjusted accordingly. To incorporate semantic and contextual adequacy of the generated text, we added another component called \em{conte}xt and seman\em{tic} based score called {\em Contanic}. It considers two fundamental expectations of a dialogue response: context relevance and adequate response, which are computed as follows:

\hspace{-0.53cm}\textbf{Context Relevance} Given a dialogue context, there may be several suitable responses. Thus, the fixed output matching approach usually suffers from a low diversity issue. Instead, assessing the relevance of a generated response in the given context and providing this feedback to the model would guide the model to generate appropriate and coherent responses. Thus, we calculate the relevance of the generated text for a context ($con$) as follows:
\begin{equation}
    CR = Cosine (e_{con}, e_{gen})
\end{equation}
\vspace{0.2em}
\begin{equation}
    \resizebox{.7\hsize}{!}{$e_{con} = BERT (<X_1, Y_1, X_2, Y_2 ... X_{t-1}, Y_{t-1}, X_t)$}
\end{equation}
\vspace{0.2em}
where $e_{con}$, and $e_{gen}$ are the representations for dialogue context and generated text, which are taken from BERT \cite{devlin2018bert}. Here, the context is compromised of all previous utterances. 

\hspace{-0.53cm}\textbf{Semantic Similarity} In natural language, we can convey the same information in various ways, i.e., a different combination of words. Thus, semantic evaluation is indeed a crucial factor in judging the adequateness of a generated text. We calculate semantic similarity (SS) between the gold sentence and the generated response as follows:
\begin{equation}
    SS = Cosine (e_{gold}, e_{gen})
\end{equation}
where $e_{gold}$ and $e_{gen}$ are semantic embedding representations for gold response and generated text, respectively. Finally, the {\em Contanic} score is computed as follows:
\begin{equation}
Contanic = \alpha \cdot CR +\beta \cdot  SS
\end{equation}
where $\alpha$ and $\beta$ are hyperparameters. The significance of dialogue context varies depending on the nature of the dialogue system. For instance, in chit-chat domains, utterances are typically less connected to the dialogue context compared to task-oriented dialogue settings. To address this discrepancy, we endeavored to construct a unified loss function applicable to both types of dialogue systems. Thus, the "Dialuation" loss function was employed for both settings, with adjustable parameters $\alpha$ and $\beta$ to accommodate the differences.\\

We experimented with the two different combinations of the CE loss and {\em Contanic}: (a) Weighted Cross-Entropy and (b) {\em Contanic} Reinforced Dialogue generation called {\em SemTextualLogue}, which are explained below.
\subsection{Weighted Cross Entropy} We first experimented with the addition of {\em Contanic} and CE loss, which performs as equivalent to CE loss. The reason was the non-differentiability of the {\em Contanic} score due to the involvement of the argmax function, and the added component becomes zero during backpropagation. To overcome this issue, we further experimented with the multiplication of these scores as a loss function. The weight parameter of the generation model is updated as follows:
\begin{equation}
\small
\begin{alignedat}{2}
w_{new} &= w_{cur} - \alpha\frac{dl_{L}}{dw}  \\
&=w_{cur}- \alpha\frac{d(1-\text{contanic})L_{CE}}{dw}  \\
&=w_{cur}-\alpha(1-\text{contanic score})\frac{dL_{CE}}{dw} 
\end{alignedat}
\end{equation}
where $L$ and $\alpha$ denote total loss and learning rate, respectively. We multiplied (1-{\em contanic}) with CE, as when gold utterance and generated text were semantically similar but lexically different, the component (1-{\em contanic}) would be lower, and the weightage to the CE loss would be reduced. Conversely, when the {\em contanic} score is low (contextually and semantically less appropriate), this component would be high, and thus the loss would be prioritized accordingly.
\subsection{{\em SemTextualLogue Loss}}
{\em SemTextualLogue} Loss aims to identify and quantify the disparities resulting from semantic, lexical, and contextual factors and then adjust the dialogue generation model's parameters based on these differences.
We build a baseline estimator, which acts as a human perception evaluator and reinforces the semantic and contextual feedback. The baseline estimator ({$BSE$}) takes the output probability distribution from the generation model and computes its relevance as follows:
\begin{equation}
    BSEscore = BSE(\hat{y})
\end{equation}
The relevance estimate is combined with the CE loss, and the final loss is computed. The loss calculation is explained below.
\begin{equation}
    L_{final}=\lambda\cdot L_{CE}+(1-\lambda)\cdot L_{SCL}+ \sigma \cdot L_{BSE}
\end{equation}
where
\begin{equation}
    L_{CE} = - \sum_{j=0}^{j=n} p(y_j) \log p (\hat{y_j})
\end{equation}

\begin{equation}
    L_{SCL} = (1-BSE score) \cdot L_{CE}
\end{equation}

\begin{equation}
    L_{BSE} = MSE (BSEscore, ContanicScore)
    \vspace{1em}
\end{equation}
where $L_{CE}$, $L_{SCL}$, and $L_{BSE}$ are the CE loss, semantic \& contextual loss, and baseline estimator loss, respectively.  The term $MSE$ indicates mean squared error loss. Here, $\lambda$ and $\sigma$ ($\lambda$, $\sigma$ $\in$ [0, 1]) are hyperparameters. In the CE loss equation, $y$ and $\hat{y}$ are true probability distributions and the predicted distribution of the gold response and the generated response, respectively. Here, $n$ is the output sequence length.

\subsection{{\em Dialution}}
\begin{figure}[hbt!]
    \centering
    \includegraphics[scale=0.67]{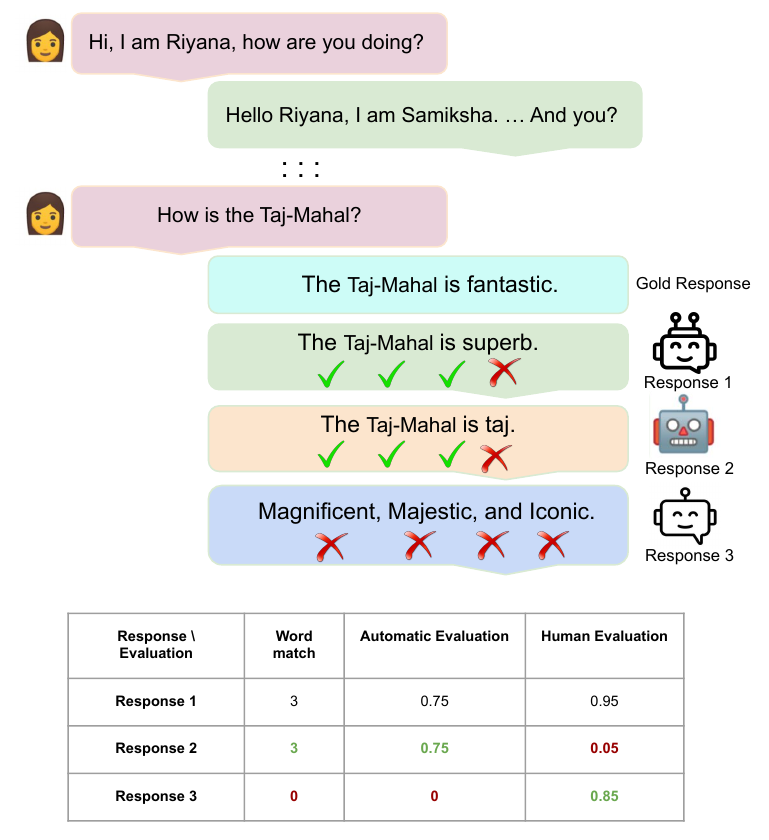}
    \vspace{-1.1em}
    \caption{One example demonstrating the significance of context and sentence semantics for evaluating dialogue responses}
        \vspace{-0.5em}
    \label{DEM}
\end{figure}

 The evaluation metrics, BLEU, ROUGE, and METEOR, only emphasize word-level matching and overlook other crucial aspects of dialogue, such as context and semantics, resulting in limited correlation with human judgments \cite{liu2016not}. One such example is illustrated in Figure \ref{DEM}. 

 {\em Response 1} is semantically very similar to the {\em gold response}, but {\em response 2} is neither meaningful nor relevant to the context. Here, the automatic and human evaluation scores are not in sync with each other. This is because the evaluation only looks at word matching, which overlooks the fact that words like ``fantastic" and ``superb" carry a similar connotation. We propose a contextualized semantic-driven dialogue evaluation metric called {\em Dialuation} to overcome the conflicts. {\em Dialuation} is a weighted average of contextual relevance (CR) and semantic score (SS). It is determined as follows:

\begin{equation}
    Dialution = (\frac{\delta_c \cdot CR + \delta_{ss} \cdot SS}{ \delta_c + \delta_{ss}}) \cdot 100
\end{equation}

 where $\delta_c$ and $\delta_{ss}$ ($\in$ [0, 1]) are the hyperparameters, which signify the importance of contextual relevance and semantic similarity, respectively. The {\em Diluation} score would lie between 0 to 100. 
 \subsection{Experimental Details}
 We experimented with the two most widely used dialogue datasets: MultiWoz 2.2 \cite{zang2020multiwoz} and PersonaChat \cite{zhang2018personalizing}. The datasets' statistics are provided in Table \ref{DS}. We employed the GPT-2 medium model and the LLaMa 7b variant for our experimentation. The train-validation-test ratios for both datasets were 8:1:1. The transformer-based model was trained for 5 epochs, while pre-trained models were trained for 3 epochs based on the convergence of loss values on an RTX 2080 Ti GPU. We have considered a context window of 3, i.e., dialogue context consists of only the last three utterances. 

\begin{table}[hbt!]
    \centering
    \vspace{-0.5em}
    \caption{Statistics of MultiWoz and PersonaChat dialogue datasets}
    \scalebox{0.95}{
    \begin{tabular}{lcc}
    \hline
    \textbf{Entity} &\textbf{MultiWoz 2.2} & \textbf{PersonaChat}  \\ \hline
      nature   & Task-oriented & Chit-Chat \\
      \# of dialogues & 9575 & 8938 \\
      \# of utterances & 71,514 & 65,719  \\
     \# of unique words & 25,714 & 18,417\\
      avg dialogue length & 7.47  & 7.35 \\
      \hline
    \end{tabular}}
    \label{DS}
    \vspace{-1em}
\end{table}
 
 The final values for hyperparameters, which are determined empirically, are as follows: source length (256), target length (256), learning rate (3e-05), batch size (32), $\alpha$ (0.3 for MultiWoz and 0.2 for PersonaChat) $\beta$ (0.7 for MultiWoz and 0.8 for PersonaChat), $\sigma$ (1), $\delta_c$ (0.3), $\delta_ss$ (0.7) and activation function (ReLU).

\section{Result and Discussion} In this section, we present the obtained experimental results, including human evaluations. Following this, we discuss the findings and analyze them in relation to the outlined research questions.

\subsection{Experimental Results}
We employed the most popular automatic evaluation metrics, namely BLEU, Rouge, and METEOR \cite{papineni2002bleu,lin2004rouge,banerjee2005meteor}, to evaluate the generation quality with different loss functions. To make the model generic, which could be applied to any dialogue setting, we have utilized only dialogue context, i.e., no additional semantic information such as intent, slot, and belief state is utilized. Thus, we compared the model with traditional CE loss and our baselines, which employ only dialogue context for response generation. 

The performances of different loss functions for different evaluation metrics on the MultiWoz and PersonaChat datasets are reported in Table \ref{R1} and Table \ref{R2}, respectively. We also measured the performances of these models in an embedding-based metric, BERT, and our newly introduced {\em Dialution} evaluation metric. The obtained results are reported in Table \ref{R3} and Table \ref{R4}. Note that all the reported values in the following tables are statistically significant, validated using the statistical t-test \cite{welch1947generalization} at a significant level of 5\%. 

\begin{table*}[hbt!]
\vspace{-0.5em}
    \centering
        \caption{Performances of various dialogue generation frameworks with different loss functions on MultiWoz dataset}
    \scalebox{0.65}{
    \begin{tabular}{lcccccccccc}
    \hline
     \textbf{Model}    & \textbf{BLEU-1} & \textbf{BLEU-2} & \textbf{BELU-3} &\textbf{BELU-4} & \textbf{BLEU}  &  \textbf{ROUGE - 1} & \textbf{ROUGE - 2} & \textbf{ROUGE- L} & \textbf{METEOR} \\ \hline
CE loss \cite{shi2021neural}   & 32.35  & 12.47 & 7.46 & 4.49 & 10.78 & 27.77 & 32.59 & 13.15 & 31.11\\
Weighted Semantic CE  & 34.16 & 13.58 & 8.27 & 5.07 & 11.55 & 27.64 & 32.38  & 13.49 & 31.02 \\
Weighted Semantic and context CE &  33.99 & 13.63 & 8.34 & 5.13 & 11.87	& 28.13 & 32.79 & 13.75 & 31.39 \\
Semantic Reinforcement  &     35.19    & 13.97     & 8.47       & 5.19        & 12.02  & 28.39  & 33.44    & 13.92  & 31.97 \\
{\em SemTextualLogue} loss  & 33.64 & 13.06 & 7.73  & 4.60  & 11.18 & 28.56 & 33.43  & 13.64   & 31.95 \\ \hline
GPT-2 w/o finetuning & 8.96 & 1.25 & 0.41 & 0.14 & 2.69 & 11.95 & 1.88 & 11.12 & 7.50  \\
GPT-2 w/ finetuning & 27.09 & 11.02 & 6.05 & 4.41 & 12.14 & 26.76 & 12.98 & 25.48 & 21.35  \\
GPT-2 w/ Weighted Semantic CE & 30.02 & 12.01 & 7.21 & 4.89 & 13.53 & 27.59 & 12.23 & 26.01 & 22.25  \\
GPT-2 w/ Weighted Semantic and context CE & 27.94 & 11.82 & 7.13 & 4.74 & 12.90 & 27.62 & 12.15 & 25.95 & 22.38  \\
GPT-2 w/ Semantic Reinforcement & 31.53 & 12.12 & 7.18 & 4.70 & 13.88 & 27.95 & 12.18 & 26.82 & 21.86  \\
GPT-2 w/ {\em SemTextualLogue} loss & 30.62&11.95 &7.02 &4.83 &13.61 &27.65 & 11.53& 26.48 & 21.73 \\  \hline
LLaMA 2 w/o finetuning &12.18 & 2.10 & 0.69 & 0.26 & 3.81 & 13.16 & 2.44 & 12.53 & 13.92 \\ 
LLaMA 2 w/ finetuning &28.22 & 12.40 & 7.55 & 5.08 & 13.31 & 27.44 & 13.63 & 26.94 & 26.47 \\ 
LLaMA 2 w/ Weighted Semantic CE & 30.79 & 14.19 & 8.97 & 5.76 & 14.93 & 28.96 & 14.77 & 27.91 & 27.70  \\
LLaMA 2 w/ Weighted Semantic and context CE & 28.53 & 13.78 & 7.92 & 4.98 & 13.80 & 27.68 & 13.82 & 27.06 & 26.33  \\
\textbf{LLaMA 2 w/ Semantic Reinforcement} & \textbf{32.29} & \textbf{14.33} & \textbf{9.05} & \textbf{5.85} & \textbf{15.38} & \textbf{29.12} & \textbf{14.83} & \textbf{27.54} & \textbf{27.90}  \\
LLaMA 2 w/ {\em SemTextualLogue} loss & 31.57 & 13.53 & 7.85  & 5.12 & 14.52 & 28.32 & 13.96 & 27.34 & 26.44 \\ 
           \hline
    \end{tabular}}
    \vspace{-1em}
    \label{R1}
\end{table*}

\begin{table*}[hbt!]
    \centering
        \caption{Performances of various dialogue generation frameworks with different loss functions on PersonaChat dataset}
    \scalebox{0.65}{
    \begin{tabular}{lcccccccccc}
    \hline
     \textbf{Model}    & \textbf{BLEU-1} & \textbf{BLEU-2} & \textbf{BELU-3} &\textbf{BELU-4} & \textbf{BLEU}  &  \textbf{ROUGE - 1} & \textbf{ROUGE - 2} & \textbf{ROUGE- L} & \textbf{METEOR} \\ \hline
    CE loss \cite{shi2021neural}  &17.72 & 3.91 & 1.08 & 0.31 & 2.21 & 17.07 & 4.41 & 16.72 & 12.73  \\ 
     Weighted Semantic CE &19.42 & 4.18 & 1.12 & 0.28 & 2.23 & 17.40 & 4.27 & 16.77 & 14.14 \\
      Weighted Semantic and Context CE &19.69 & 4.19 & 1.16 & 0.37 & 2.30 & 17.47 & 4.40 & 17.15 & 12.96 \\
      Semantic Reinforcement &18.85 & 4.33 & 1.16 & 0.33 & 2.36 & 15.96 & 4.18 & 15.63 & 13.52 \\
     {\em SemTextualLogue} loss & 20.17 & 4.41 & 1.19 & 0.36 & 2.37 & 17.44 & 4.37 & 17.09 & 13.26  \\ \hline
      GPT-2 w/o finetunning & 6.25 & 1.05 & 0.22 & 0.05 & 1.89 & 13.95 & 2.84 & 13.58 & 8.68  \\ 
     GPT-2 w/ finetuning & 23.12 & 6.11 & 2.03 & 0.74 & 7.99 & 18.71 & 5.20 & 17.98 & 11.59  \\ 
    GPT-2 w/ Weighted Semantic CE & 24.23 & 6.89 & 2.58 & 0.92 & 8.66 & 17.59 & 5.12 & 18.05 & 11.68  \\ 
    GPT-2 w/ Weighted Semantic \& Context CE & 23.85 & 6.05 & 1.95 & 0.83 & 8.17 & 17.43 & 4.97 & 17.53 & 11.32  \\ 
    GPT-2 w/ Semantic Reinforcement  & 24.56 & 6.91 & 2.64 & 0.98 & 8.77 & 17.63 & 5.23 & 17.83 & 11.74  \\
    GPT-2 w/ {\em SemTextualLogue} loss & 24.09 & 6.23 & 1.87 & 0.74 & 8.23 & 17.02 & 5.07 & 17.95 & 11.61 \\ \hline
    LLaMA 2 w/o finetuning & 5.97 & 0.47 & 0.10 & 0.02 & 1.64 & 7.92 & 0.68 & 7.23 & 11.36  \\ 
    LLaMA 2 w/ finetuning & 24.18 & 6.56 & 2.66 & 0.98 & 8.60 & 19.12 & 5.78 & 18.50 & 12.26 \\ 
    LLaMA 2 w/ Weighted Semantic CE & 26.44 & 7.27 & 2.73 & 1.08 & 9.37  & 20.44 & 6.74 & 18.72 & 13.64 \\ 
    LLaMA 2 w/ Weighted Semantic \& Context CE & 25.53 & 6.91 &2.53 & 0.85& 8.96& 19.85&6.31 & 18.38 & 12.69 \\ 
   \textbf{LLaMA 2 w/ Semantic Reinforcement} & \textbf{26.69} & \textbf{7.31} & \textbf{2.84} & \textbf{0.95} & \textbf{9.45} & \textbf{20.39} & \textbf{6.81}& \textbf{18.74} & \textbf{13.72} \\ 
    LLaMA 2 w/ {\em SemTextualLogue} loss & 25.67 & 6.83 & 2.58 & 0.79 & 8.97 & 19.54&6.37 & 18.49 & 12.78 \\ 
           \hline
    \end{tabular}}
    \label{R2}
    \vspace{0.5em}
\end{table*}
\begin{table}[hbt!]
    \begin{minipage}{0.47\textwidth}
    \centering
    \caption{Vector-embedding based evaluation result on Multiwoz dataset}
    \scalebox{0.63}{
    \begin{tabular}{lcc}
    \hline
      \textbf{Model}   & \textbf{BERT Score} & \textbf{Dialution}  \\ \hline
     CE \cite{shi2021neural}  & 57.58 & 51.43 \\ 
      Weighted Semantic CE & 57.91 & 51.22 \\ 
      Weighted Semantic and Context CE & 57.94 & 51.27 \\
      Semantic Reinforcement & 58.36 & 51.85  \\
      {\em SemTextualLogue} loss & 58.83 & 52.38 \\
      GPT-2 w/o finetuning & 43.43 & 46.46 \\
      GPT-2 w/ finetuning & 60.80 & 55.58 \\
      GPT-2 w/ Weighted Semantic CE & 62.58 & 57.49 \\
      GPT-2 w/ Weighted Semantic \& Context CE & 61.05 & 60.69 \\
      GPT-2 w/ Semantic Reinforcement & 63.24 & 60.56 \\
      GPT-2 w/ {\em SemTextualLogue} loss & 63.39 & 61.83 \\
      LLaMA 2 w/o finetuning & 45.98 & 48.96 \\
      LLaMA 2 w/ finetuning & 63.92 & 57.61 \\
      LLaMA 2 w/ Weighted Semantic & 64.57 & 58.11 \\
      LLaMA 2 w/ Weighted Semantic \& Context CE & 61.52 & 59.63 \\
      LLaMA 2 w/ Semantic Reinforcement & 64.98 & 61.11 \\
      \textbf{LLaMA 2 w/ {\em SemTextualLogue} loss} & \textbf{65.38} & \textbf{62.62} \\
      \hline
    \end{tabular}}
    \label{R3}
    \end{minipage}%
    \hspace{1em}
    \begin{minipage}{0.47\textwidth}
    \centering
    \caption{Vector-embedding based evaluation result on PersonaChat dataset}
    \scalebox{0.63}{
    \begin{tabular}{lcc}
    \hline
      \textbf{Model}   & \textbf{BERT Score} & \textbf{Dialution}  \\ \hline
     CE \cite{shi2021neural}  & 32.72 & 26.64 \\ 
      Weighted Semantic CE & 32.84 & 28.50\\ 
      Weighted Semantic and Context CE & 34.05 & 28.82 \\
      Semantic Reinforcement & 33.35 & 26.52 \\
      {\em SemTextualLogue} loss & 34.37 & 29.94 \\
      GPT-2 w/o finetuning & 35.53 & 38.62 \\
      GPT-2 w/ fintuning & 41.39 & 41.20 \\
     GPT-2 w/ Weighted Semantic CE & 42.03 & 41.28 \\
      GPT-2 w/ Weighted Semantic \& Context CE & 40.23 & 41.39 \\
      GPT-2 w/ Semantic Reinforcement & 42.37 & 41.61 \\
      GPT-2 w/ {\em SemTextualLogue} loss & 42.78 & 42.96 \\
      LLaMA 2 w/o finetuning & 38.71 & 41.61 \\
      LLaMA 2 w/ finetuning & 41.92 & 41.66 \\
      LLaMA 2 w/ Weighted Semantic CE & 42.83 & 41.76 \\
      LLaMA 2 w/ Weighted Semantic \& Context CE & 41.25 & 42.18 \\
      LLaMA 2 w/ Semantic Reinforcement & 43.08 & 42.31 \\
     \textbf{LLaMA 2 w/ {\em SemTextualLogue} loss} & \textbf{43.82} & \textbf{44.13} \\
      \hline
    \end{tabular}}
    \label{R4}
    \end{minipage}
    \vspace{-0.5em}
\end{table}

\hspace{-0.53cm}\textbf{Human Evaluation} We also conducted the human evaluation of 150 test samples from each dataset. In this assessment, three researchers (other than the authors) were employed to evaluate the generated responses (50 samples of each model) of different models without revealing their names. The samples are assessed based on the following five metrics: \textit{adequacy, fluency, coherence, naturalness, and completeness} on a scale of 0 to 5. The obtained scores for both datasets are provided in Table \ref{HE_M} and Table \ref{HE_P}. The findings show a similar trend as we found in the automated evaluation results.

\begin{table}[hbt!]
    \centering
    \vspace{-0.5em}
    \caption{Human evaluation of LLaMa 2 models with different loss functions on MultiWoz dataset}
   \scalebox{0.72}{
\begin{tabular}{lcccccc}
\hline
\textbf{Model} & \textbf{Adequacy} & \textbf{Fluency} & \textbf{Context Relevance} & \textbf{Naturalness} & \textbf{Completeness} & \textbf{Avg.} \\ \hline
CE & 3.86 & 4.40 & 4.10 & 3.64 & 3.78 & 3.96 \\
Weighted Semantic CE & 3.94 & 4.42 & 4.18 & 3.60 & 3.70 & 3.97 \\
Weighted Semantic and Context CE & 4.12 & \textbf{4.48} & 4.32 & 3.73 & 3.96 & 41.2 \\
Semantic Reinforcement & 4.38 & 4.46 & 4.28 & 3.68 & 4.12 & 4.18 \\
\textbf{{\em SemTextualLogue} loss} & \textbf{4.42} & 4.47 & \textbf{4.36} & \textbf{3.77} & \textbf{4.26} & \textbf{4.26} \\
\hline
\end{tabular}}
\label{HE_M}
\vspace{-3em}
\end{table}

\begin{table}[hbt!]
\centering
\caption{Human evaluation of LLaMa 2 models with different loss functions on PersonaChat dataset}
   \scalebox{0.72}{
\begin{tabular}{lcccccc}
\hline
\textbf{Model} & \textbf{Adequacy} & \textbf{Fluency} & \textbf{Context Relevance} & \textbf{Naturalness} & \textbf{Completeness} & \textbf{Avg.} \\ \hline
CE  & 2.16 & 3.08 & 3.81 & 4.02 & 2.26 & 3.07 \\
Weighted Semantic CE & 2.30 & 3.36 & 4.06 & 4.28 & 2.38 & 3.07 \\
Weighted Semantic and Context CE & 2.30 & 3.36 & 4.06 & 4.28 & 2.38 & 3.28 \\
Semantic Reinforcement & 2.33 & 3.38 & 4.11 & 4.34 & 2.44 & 3.32 \\
\textbf{{\em SemTextualLogue} loss} & \textbf{2.38} & \textbf{3.44} & \textbf{4.16} & \textbf{4.45} & \textbf{2.48} & \textbf{3.38} \\
\hline
\end{tabular}}
\vspace{-1.7em}
\label{HE_P}
\end{table}

\subsection{Findings and Observations} Based on the experimental findings, we report the following answers (with evidence) to our investigated research questions (RQs).\\

\hspace{-0.53cm}\textbf{RQ 1: Can adding a semantic-based evaluation component to the lexical-based loss function provide more accurate feedback on generated responses and thus improve dialogue generation quality?}\\
The performances of models with traditional CE  and proposed Semantic Reinforcement and {\em SemTextualLogue} loss functions are reported in Table \ref{R1}, Table \ref{R2} (in terms of traditional evaluation metrics), Table \ref{R3}, and Table \ref{R4} (in terms of BERT and Dialuation scores).  We can see a significant improvement across different base models and evaluation metrics on the datasets (CE vs Weighted Semantic CE and CE vs Semantic Reinforcement): MultiWoz (BERT: +1.46, {\em Dialution}: 5.01, BLEU: +2.07, ROUGE-L: +0.60, and METEOR: +1.43), PersonaChat (BERT Score: +1.90, {\em Dialution}: +2.47, BLEU: +0.85, ROUGE-L: +0.24, and METEOR: +1.46). Moreover, we also observed a significant enhancement in human evaluation. These improvements firmly establish that there is a role of semantic evaluation infusion in the loss function. 

\hspace{-0.53cm}\textbf{RQ 2: Can incorporating context relevance evaluation in the loss function improve the model's ability to generate more appropriate and coherent responses to the discourse?}\\
We did not observe significant improvement in lexical metrics, but we observed a consistent improvement pattern across the embedding-based evaluation metrics when we utilized dialogue context relevance in the loss function modeling (Table \ref{R3}: Semantic Reinforcement vs. {\em SemTextualLogue} Loss and Table \ref{R4}: Semantic Reinforcement vs. {\em SemTextualLogue} Loss). Similar behavior has also been found in human evaluations, which are reported in Table \ref{HE_M} and Table \ref{HE_P}.  We believe the pattern arises from the multi-objective expectation, seeking alignment with the dialogue context and the gold response. Consequently, the system is discovering an optimal balance, consistently demonstrating improved performance in embedding metrics and human evaluation. The inclusion of context provides additional feedback to the dialogue generation model about the adequateness of the generated response, leading to contextualized responses. 

\hspace{-0.53cm}\textbf{RQ 3: Will integrating the semantic and contextual components to the lexical-based evaluation metrics in dialogue generation result in better correlation with human judgment?}\\
We found that the {\em SemTexualLogue} performance on the Multiwoz dataset is very close to baselines in terms of BLEU; however, the model significantly outperforms others in human evaluation. A similar notation as human evaluation is being reflected in our newly introduced loss function, {\em Dialuation}. We found many cases where a response was very relevant but did not match with the gold utterance; thus, both {\em Dialuation} and human score were high despite low BLEU score. One such instance is as follows: {\em context}: Hi, ...lets watch a new movie, {\em generated:} I prefer some new web-series, {\em gold:} lets go! We can watch it. It firmly shows that the {\em Diluation}, which considers both semantic and context relevance, is more aligned with human evaluation than any other metrics, including the BERT score.

\hspace{-0.53cm}\textbf{Key Observations} The experiments and analysis yield key observations that are worth noting, as follows: (a) Both GPT and LLaMMa (w/o finetuning) fell short of surpassing the performance of a transformer-based model that had undergone fine-tuning. This suggests the challenge of capturing diverse behavior within a local context without additional training. (b) We noticed that models incorporating both dialogue context and semantics outperform those using only Cross-Entropy (CE) loss but fall short in terms of lexical metrics compared to models with only the semantic component. However, these context-aware models excel in both BERT score and {\em Dialution}, indicating that contextual information contributes to generating relevant responses, even if they may not precisely match the lexical patterns of the corpus behavior. (c) We note that the scoring scale for MultiWoz is comparatively higher than that of PersonaChat. This difference can be attributed to the diversity and elevated entropy present in the open-domain dataset, PersonaChat. 

\section{Case Study and Analysis} We have analyzed the models' performances for common test cases, and a few samples are shown in Table \ref{CS}. The comprehensive analyses of the performances of different models lead to the following key observations: \textbf{(i)}  Due to the incorporation of semantics and context, our model generates more contextualized and consistent responses, as shown in Table \ref{CS}. \textbf{(ii)} We observed that the number of synonyms for a word is very limited in the MultiWoz dataset's vocabulary space, so the influence of semantic infusion is comparatively less than the performance gain observed on the PersonaChat dataset. \textbf{(iii)} In some cases, more often in the persona dataset, the proposed model repeats some words (persona entities) in response, primarily due to the fact that they occur more often in the corpora. 

\begin{table}[hbt!]
    \centering
    \caption{Case study- generated responses by LLaMa 2 with different variants of loss functions for some common test cases}
    \scalebox{0.82}{
    \begin{tabular}{p{14cm}}
         \hline
         \textbf{Context:} Context: $<domain>$train, hotel $<domain>$$<history>$ STARTOFDIALOGUE $<history>$ \\
         \textbf{Utterance:}  I am looking for a place to stay in the north of the city. I would prefer a 4 star hotel please.\\
         \textbf{Gold Response:} There are several guesthouses available. Do you have a price reference? \\
         \textbf{CE:} I have two hotels in the north and one in the moderate price range. Do you have a preference?  \\ 
          \textbf{Weighted Semantic and Context CE:}  I have two hotels that meet your criteria. One is in the north and the other is in the moderate price range. Do you have a preference? <eos>\\ 
         \textbf{SemTextual Logue:} There are two hotels that meet your criteria. The Ashley Hotel and the Lovell Lodge. Would you like to book one of these?
         \\ \hline
         \textbf{Context:} $<domain>$ restaurant, train $<domain>$$<history>$ $<u>$ I am looking for a train departing from Cambridge, leaving on Tuesday. $<u>$............$<s>$ I found 3 places, would you like to try The Cow Pizza Kitchen and Bar? $<s>$$<u>$ Yes, please book a reservation for 7 people. $<u>$$<s>$Is there anything else I can do for you? $<s>$$<u>$ Can I get the reference number for that. $<u>$$<s>$ What day and time would you like the reservation for? $<s>$\\
         \textbf{Utterance:} Tuesday at 10:00 \\
         \textbf{Gold Response:} I was able to book that for you. Your reference number is RRRRR. Your table will be reserved for 15 minutes. \\
         \textbf{CE:} I have booked your table for 7. \\
         \textbf{Weighted Semantic and Context CE:} I have booked your table at the Cow pizza kitech and bar  \\ 
         \textbf{SemTextual Logue:} I have booked your table for 7 at the Cow pizza kitech and bar. Your reference number is: RRRRR. What else can I help you with?\\ 
         \hline
    \end{tabular}}
    \label{CS}
    \vspace{-0.5em}
\end{table}

Our introduced loss function and evaluation metrics exhibit superior performance over traditional counterparts. However, it is crucial to acknowledge inherent limitations. The limitations are summarized as follows: (i) The coherence between a response and the dialogue history is influenced by the nature of the dialogue. In task-oriented dialogues, each response is typically more closely aligned with the context compared to chit-chat conversations. Consequently, the coefficient ($\alpha$) for incorporating context in the loss function varies accordingly. As a result, we identified two different values that proved effective for these two settings. However, the optimal coefficient values for these two different natures of dialogue can be determined using experimentation with various datasets. (ii) We observed that despite incorporating a contextual semantic vector for sentence representation, it struggles to capture the similarity between two sentences when one of them contains a negation with an antonym of a word contained in the other sentence, even if the overall meaning is similar. (iii) The dialogue generation framework we propose, incorporating the {\em SemTextualLogue} requires slightly more time for training compared to models trained with Cross-Entropy (CE) loss. The model with the CE loss function takes 5 hours whereas {\em SemTextualLogue}Loss takes 6.30 hours for 1 epoch. It is crucial to highlight that there is no variance in inference times between the two.

\section{Conclusion}
The core of a learning framework is the objective function and evaluation metrics, which are used to train the underlying task and assess its performance. Cross entropy (CE) and BLEU are the commonly employed loss function and evaluation metric for dialogue generation, but they have the drawback of fixed target comparison. We propose a semantic-infused contextualized dialogue ({\em SemTextual Logue}) loss function to address this issue. Moreover, we introduced a new dialogue evaluation metric called {\em Dialuation}, which also considers dialogue context in addition to gold text to assess the relevance of the generated response. We experimented with both kinds of dialogue corpora, namely task-oriented and chit-chat. The proposed {\em SemTextual Logue} obtained superior performance on both datasets across various evaluation metrics, including human evaluation. The obtained improvements and analysis firmly establish the efficacy of dialogue context and semantic evaluation for dialogue generation loss function. When we evaluate a response, we implicitly use global knowledge in addition to the context, and thus, an evaluation by a child and an evaluation by an experienced individual differ. In the future, we would like to investigate the role of external knowledge in developing an appropriate loss function.

\section{Ethical Consideration} Ethics play a foundational role in research and development efforts, and therefore, we considered ethical considerations in every stage of our research process, spanning from hypothesis formulation to analysis. We have conscientiously integrated ethical considerations into our work, utilizing two widely recognized benchmark datasets for dialogue system research. Furthermore, we have made both the datasets and code publicly available on an anonymous GitHub repository, facilitating progress in dialogue generation loss function research. To ensure impartiality in human evaluation, we engaged three annotators who assessed generated samples without knowledge of the underlying models, rating them based on predefined metrics. Through these measures, we uphold ethical guidelines and principles, promoting transparency and integrity in our research endeavors.

\begin{thebibliography}{10}
\providecommand{\url}[1]{\texttt{#1}}
\providecommand{\urlprefix}{URL }
\providecommand{\doi}[1]{https://doi.org/#1}

\bibitem{allen2001toward}
Allen, J.F., Byron, D.K., Dzikovska, M., Ferguson, G., Galescu, L., Stent, A.: Toward conversational human-computer interaction. AI magazine  \textbf{22}(4),  27--27 (2001)

\bibitem{banerjee2005meteor}
Banerjee, S., Lavie, A.: Meteor: An automatic metric for mt evaluation with improved correlation with human judgments. In: Proceedings of the acl workshop on intrinsic and extrinsic evaluation measures for machine translation and/or summarization. pp. 65--72 (2005)

\bibitem{conley2018improving}
Conley, T., Clair, J.S., Kalita, J.: Improving computer generated dialog with auxiliary loss functions and custom evaluation metrics. In: 15th International Conference on Natural Language Processing. p.~143 (2018)

\bibitem{de2005tutorial}
De~Boer, P.T., Kroese, D.P., Mannor, S., Rubinstein, R.Y.: A tutorial on the cross-entropy method. Annals of operations research  \textbf{134},  19--67 (2005)

\bibitem{devlin2018bert}
Devlin, J., Chang, M.W., Lee, K., Toutanova, K.: Bert: Pre-training of deep bidirectional transformers for language understanding. arXiv preprint arXiv:1810.04805  (2018)

\bibitem{feng2021survey}
Feng, X., Feng, X., Qin, B.: A survey on dialogue summarization: Recent advances and new frontiers. arXiv preprint arXiv:2107.03175  (2021)

\bibitem{golovanov2019large}
Golovanov, S., Kurbanov, R., Nikolenko, S., Truskovskyi, K., Tselousov, A., Wolf, T.: Large-scale transfer learning for natural language generation. In: Proceedings of the 57th Annual Meeting of the Association for Computational Linguistics. pp. 6053--6058 (2019)

\bibitem{griol2008statistical}
Griol, D., Hurtado, L.F., Segarra, E., Sanchis, E.: A statistical approach to spoken dialog systems design and evaluation. Speech Communication  \textbf{50}(8-9),  666--682 (2008)

\bibitem{hu2020translation}
Hu, W., Le, R., Liu, B., Ma, J., Zhao, D., Yan, R.: Translation vs. dialogue: A comparative analysis of sequence-to-sequence modeling. In: Proceedings of the 28th International Conference on Computational Linguistics. pp. 4111--4122 (2020)

\bibitem{jiang2019improving}
Jiang, S., Ren, P., Monz, C., de~Rijke, M.: Improving neural response diversity with frequency-aware cross-entropy loss. In: The World Wide Web Conference. pp. 2879--2885 (2019)

\bibitem{kovaleva2018similarity}
Kovaleva, O., Rumshisky, A., Romanov, A.: Similarity-based reconstruction loss for meaning representation. In: Proceedings of the 2018 Conference on Empirical Methods in Natural Language Processing. pp. 4875--4880 (2018)

\bibitem{li2017adversarial}
Li, J., Monroe, W., Shi, T., Jean, S., Ritter, A., Jurafsky, D.: Adversarial learning for neural dialogue generation. In: EMNLP (2017)

\bibitem{lin2004rouge}
Lin, C.Y.: Rouge: A package for automatic evaluation of summaries. In: Text summarization branches out. pp. 74--81 (2004)

\bibitem{liu2016not}
Liu, C.W., Lowe, R., Serban, I.V., Noseworthy, M., Charlin, L., Pineau, J.: How not to evaluate your dialogue system: An empirical study of unsupervised evaluation metrics for dialogue response generation. In: Proceedings of the 2016 Conference on Empirical Methods in Natural Language Processing. pp. 2122--2132 (2016)

\bibitem{nakamura2019another}
Nakamura, R., Sudoh, K., Yoshino, K., Nakamura, S.: Another diversity-promoting objective function for neural dialogue generation  (2019)

\bibitem{ni2023recent}
Ni, J., Young, T., Pandelea, V., Xue, F., Cambria, E.: Recent advances in deep learning based dialogue systems: A systematic survey. Artificial intelligence review  \textbf{56}(4),  3055--3155 (2023)

\bibitem{niwattanakul2013using}
Niwattanakul, S., Singthongchai, J., Naenudorn, E., Wanapu, S.: Using of jaccard coefficient for keywords similarity. In: Proceedings of the International MultiConference of Engineers and Computer Scientists. vol.~1 (2013)

\bibitem{papineni2002bleu}
Papineni, K., Roukos, S., Ward, T., Zhu, W.J.: Bleu: a method for automatic evaluation of machine translation. In: Proceedings of the 40th annual meeting of the Association for Computational Linguistics. pp. 311--318 (2002)

\bibitem{radford2019language}
Radford, A., Wu, J., Child, R., Luan, D., Amodei, D., Sutskever, I., et~al.: Language models are unsupervised multitask learners. OpenAI blog  \textbf{1}(8), ~9 (2019)

\bibitem{sato2020evaluating}
Sato, S., Akama, R., Ouchi, H., Suzuki, J., Inui, K.: Evaluating dialogue generation systems via response selection. In: Proceedings of the 58th Annual Meeting of the Association for Computational Linguistics. pp. 593--599 (2020)

\bibitem{sellam2020bleurt}
Sellam, T., Das, D., Parikh, A.: Bleurt: Learning robust metrics for text generation. In: Proceedings of the 58th Annual Meeting of the Association for Computational Linguistics. pp. 7881--7892 (2020)

\bibitem{serban2016building}
Serban, I., Sordoni, A., Bengio, Y., Courville, A., Pineau, J.: Building end-to-end dialogue systems using generative hierarchical neural network models. In: Proceedings of the AAAI conference on artificial intelligence. vol.~30 (2016)

\bibitem{shen2021text}
Shen, L., Zhan, H., Shen, X., Song, Y., Zhao, X.: Text is not enough: Integrating visual impressions into open-domain dialogue generation. In: Proceedings of the 29th ACM International Conference on Multimedia. pp. 4287--4296 (2021)

\bibitem{shi2021neural}
Shi, T., Keneshloo, Y., Ramakrishnan, N., Reddy, C.K.: Neural abstractive text summarization with sequence-to-sequence models. ACM Transactions on Data Science  \textbf{2}(1),  1--37 (2021)

\bibitem{touvron2023llama}
Touvron, H., Lavril, T., Izacard, G., Martinet, X., Lachaux, M.A., Lacroix, T., Rozi{\`e}re, B., Goyal, N., Hambro, E., Azhar, F., et~al.: Llama: Open and efficient foundation language models. arXiv preprint arXiv:2302.13971  (2023)

\bibitem{ueyama2020diverse}
Ueyama, A., Kano, Y.: Diverse dialogue generation with context dependent dynamic loss function. In: Proceedings of the 28th International Conference on Computational Linguistics. pp. 4123--4127 (2020)

\bibitem{valizadeh2022ai}
Valizadeh, M., Parde, N.: The ai doctor is in: A survey of task-oriented dialogue systems for healthcare applications. In: Proceedings of the 60th Annual Meeting of the Association for Computational Linguistics (Volume 1: Long Papers). pp. 6638--6660 (2022)

\bibitem{wang2021diversifying}
Wang, Y., Zheng, Y., Jiang, Y., Huang, M.: Diversifying dialog generation via adaptive label smoothing. In: Proceedings of the 59th Annual Meeting of the Association for Computational Linguistics and the 11th International Joint Conference on Natural Language Processing (Volume 1: Long Papers). pp. 3507--3520 (2021)

\bibitem{welch1947generalization}
Welch, B.L.: The generalization ofstudent's' problem when several different population variances are involved. Biometrika  \textbf{34}(1/2),  28--35 (1947)

\bibitem{zang2020multiwoz}
Zang, X., Rastogi, A., Sunkara, S., Gupta, R., Zhang, J., Chen, J.: Multiwoz 2.2: A dialogue dataset with additional annotation corrections and state tracking baselines. ACL 2020 p.~109 (2020)

\bibitem{zhang2018personalizing}
Zhang, S., Dinan, E., Urbanek, J., Szlam, A., Kiela, D., Weston, J.: Personalizing dialogue agents: I have a dog, do you have pets too? In: Proceedings of the 56th Annual Meeting of the Association for Computational Linguistics (Volume 1: Long Papers). pp. 2204--2213 (2018)

\bibitem{zhang2019bertscore}
Zhang, T., Kishore, V., Wu, F., Weinberger, K.Q., Artzi, Y.: Bertscore: Evaluating text generation with bert. In: International Conference on Learning Representations (2019)

\bibitem{zhao2020knowledge}
Zhao, X., Wu, W., Xu, C., Tao, C., Zhao, D., Yan, R.: Knowledge-grounded dialogue generation with pre-trained language models. In: Proceedings of the 2020 Conference on Empirical Methods in Natural Language Processing (EMNLP). pp. 3377--3390 (2020)

\end{thebibliography}

\end{document}